\documentclass{article}

\usepackage[nonatbib,final]{neurips_2024}

\usepackage[utf8]{inputenc} 
\usepackage[T1]{fontenc}    
\usepackage{hyperref}       
\usepackage{url}            
\usepackage{booktabs}       
\usepackage{amsfonts}       
\usepackage{nicefrac}       
\usepackage{microtype}      
\usepackage{xcolor}         

\usepackage{multirow}
\newcommand{\specialcell}[2][l]{%
  \begin{tabular}[#1]{@{}l@{}}#2\end{tabular}}
\usepackage[acronym]{glossaries}
\newacronym{MAE}{MAE}{mean absolute error}
\newacronym{VC}{VC}{VC}
\newacronym{GT}{ground truth}{ground truth}
\usepackage{amsmath,amssymb}
\usepackage{mathtools}
\usepackage{relsize}
\usepackage{tipa}
\usepackage{textcomp}
\usepackage{hyperref}

\usepackage{svg}
\usepackage{siunitx}
\usepackage{enumitem}
\usepackage[subtle]{savetrees}

\title{Improving Voice Quality in Speech Anonymization With Just Perception-Informed Losses}

\author{%
  Suhita Ghosh*,
  Tim Thiele*,
  Frederic Lorbeer*,
  Frank Dreyer*,
  Sebastian Stober\\
  Artificial Intelligence Lab (AILab), Otto-von-Guericke-University, Magdeburg, Germany\\
  \texttt{\{suhita.ghosh, stober\}@ovgu.de}
}

\begin{document}
\def\thefootnote{*}\footnotetext{These authors contributed equally to this work.}\def\thefootnote{\arabic{footnote}}

\maketitle

\begin{abstract}
The increasing use of cloud-based speech assistants has heightened the need for effective speech anonymization, which aims to obscure a speaker's identity while retaining critical information for subsequent tasks. One approach to achieving this is through voice conversion. While existing methods often emphasize complex architectures and training techniques, our research underscores the importance of loss functions inspired by the human auditory system. Our proposed loss functions are model-agnostic, incorporating handcrafted and deep learning-based features to effectively capture representations about speech quality. Through objective and subjective evaluations, we demonstrate that a VQVAE-based model, enhanced with our perception-driven losses, surpasses the vanilla model in terms of naturalness, intelligibility, and prosody while maintaining speaker anonymity. These improvements are consistently observed across various datasets, languages, target speakers, and genders.
\end{abstract}

\section{Introduction}
Over the last few years, cloud-based speech devices, such as voice assistants, have become indispensable in life~\cite{zwakman2021usability}.
However, this also poses increasing privacy threats, as speech data contains sensitive information encompassing health, affiliations, and other private information about the speaker~\cite{IntroI_Trust_VoiceAssistants, HaaseHCII2022}. 
Therefore, speech anonymization becomes pertinent, which hides the personal identifiers in the speech while retaining the linguistic content.
Voice conversion (VC) is one of the ways to achieve speech anonymization, where the source utterance is modified to sound like another `target' speaker.
In cases where the response of a
speech device is driven by the end-user's emotional state, preserving prosody becomes crucial, such as in health monitoring systems that adjust alert urgency based on detected stress or anxiety to ensure timely intervention.

Research on statistical modelling approaches~\cite{takashima2013exemplar,jin2016cute,helander2011voice,sun2016phonetic} provided the groundwork for the development of deep learning (DL)-based VC techniques, significantly advancing the state-of-the-art in VC research~\cite{walczyna2023overview,popovdiffusion}.
Most of the VC methods are based on generative adversarial networks (GANs)~\cite{sisman2020overview}, which produce natural-sounding conversions.
This is due to the discriminator's role in guiding the generator to create conversions consistent with the target speaker's characteristics.
Current GAN-based VC methods~\cite{das2023stargan, EmoStarGAN} are trained with over seven losses in addition to adversarial losses, complicating training due to instability in the optimization process and heightened sensitivity to hyperparameter choices~\cite{GANs_disadvantage}.
In contrast to GANs, variational autoencoders (VAEs)~\cite{williams2021learning,georges2022self} offer a clear advantage by providing a well-defined likelihood function, ensuring a more stable training than GANs.
These methods typically disentangle the speaker and content embeddings using a reconstruction loss and relevant constraints to remove speaker information.

One notable variant of VAE, the vector-quantized VAE (VQVAE)~\cite{VQVAE}, uses a discrete distribution over a codebook instead of a continuous distribution, which is potentially a more intuitive approach given that language is inherently discrete, such as speech being represented as a sequence of phonemes.
Recent VQVAE-based approaches organize the latent embeddings by the phonetic content~\cite{Ding2019GroupLE} or arrange them in a hierarchical manner~\cite{HLE-VQVAE}, to capture the different semantic levels of speech across various temporal scales.
However, VQVAE-based approaches have garnered limited traction compared to GANs, due to their tendency to produce averaged outputs, leading to a buzzy-sounding voice~\cite{walczyna2023overview}.
One reason for this issue is that VQVAEs typically use an element-wise loss function in the output space~\cite{pixel_wise_problem}.
These losses do not penalize the regions which are pertinent to the human auditory system~\cite{CODEC_Network}.
This leads to low-quality reconstructions with dampened prosody.

Thus, we propose novel `perception-informed' losses to enhance the quality of speech produced by a VC model.
These losses aim to introduce an inductive bias that may aid in achieving higher fidelity reconstructions.
We propose two kinds of losses: handcrafted feature-based and representation-driven.
The handcrafted feature-based loss is computed on formants, which represent the resonance frequencies of the vocal tract and are crucial for defining the characteristics of vowel sounds, playing a significant role in speech perception and identifying phonetic elements.
Drawing from perceptual losses in audio enhancement tasks~\cite{denoising_withFeatureLoss}, our representation-driven losses emphasize the aspects of sound most critical to human listeners, thereby improving the perceptual fidelity of the generated speech.
Although our proposed loss functions can be integrated into any model, we consider a VQVAE-based model to demonstrate their effectiveness, as training a VQVAE model is generally simpler than training a GAN-based model.
We extensively evaluate our approach using various datasets, languages, target speakers, and genders. We demonstrate through objective and subjective tests that our proposed loss functions generalize well and significantly enhance speech quality across different scenarios.

\section{Vanilla VQVAE} \label{section:baseline}
VQVAE achieves VC by transforming the source mel-spectrogram $x$ using the speaker embedding of the target speaker, typically learned during training~\cite{HLE-VQVAE}. There are three key components of VQVAE:
\begin{enumerate}[leftmargin=0.7cm,itemsep=0mm,labelsep=0.1cm]
\item \textbf{Encoder} $Enc$ takes a mel-spectrogram $x$ and maps it to a discrete latent variable $z = Enc(x)$, which is received by the vector quantization layer.
\item \textbf{Vector Quantization}
layer, also known as the codebook $C_c$, sits between the encoder $Enc$ and the decoder $Dec$.
This layer consists of learnable vectors representing embeddings that capture speaker-independent content information. The encoder's output $z$ is used to select the most similar vector $q$ from this codebook based on Euclidean distance. This selected vector $q$ is then passed to the decoder in place of $z$. Since the nearest vector selection is non-differentiable, the straight-through re-parameterization trick is applied to compute the discrete latent vector $q_{st}$, as $q_{st} = z + sg(q - z)$, where $sg$ is the stop-gradient operator~\cite{VQVAE}.

\item \textbf{Decoder} $Dec$ receives two inputs: the content embedding $q_{st}$ and a speaker embedding $e_s$, selected from speaker codebook $C_s= \{e_s\}_{s=1}^S,  s\in1...S$.
The speaker codebook $C_s$ is jointly optimized with the other model parameters during training through back-propagation.
Using both of these inputs $q_{st}$ and $e_s$, the decoder generates the transformed mel-spectrogram $x_{dec} = Dec(q_{st}, e_s)$.
Therefore, VC using VQVAE can be achieved by just replacing the source speaker embedding with the target speaker embedding.
\end{enumerate}



We use a hierarchical-based VQVAE~\cite{HLE-VQVAE} as our baseline, which employs $L$=3 levels of vector quantization layers to capture speech representations at varying semantic depths (e.g., phoneme, syllable, word), enhancing reconstruction quality. The model is trained with the loss functions shown in Equation~\ref{eq:baseline_loss}: reconstruction loss for preserving linguistic content, codebook loss to ensure that the encoded representations remain close to the discrete codebook vectors, and commitment loss ensures latent representations remain consistent with specific codebook vectors~\cite{VQVAE}. Each loss is weighed by hyperparameter $\lambda$.
\begin{equation}
\label{eq:baseline_loss}
    \begin{aligned}
         L_{\text{vanilla}} &= \lambda_{\text{recon}}  \lVert x - x_{dec} \rVert_{2}^{2}
         + \lambda_{\text{code}} \sum_{l=1}^{L} \lVert sg [z_{l}] - q_{l} \rVert_{2}^{2}
         + \lambda_{\text{com}} \sum_{l=1}^{L} \lVert z_{l} - sg [q_{l}] \rVert_{2}^{2}
    \end{aligned}
\end{equation}

\section{Perception-Informed Losses}
Recent VC research has mainly focused on enhancing architectures to improve synthesized speech quality, often leading to complex models and overfitting~\cite{walczyna2023overview}. In contrast, our approach introduces novel loss functions, applicable to any model, that aim to capture speech quality in line with human perception.


\textbf{Handcrafted Feature-Based Loss:} 
Formants serve as a concise descriptor of the spectral content of vowels, efficiently capturing important speech features with minimal parameters~\cite{SpringerHandbook_SpeechProcessing}.
Phonetically, formants are resonant frequencies that are characteristic of the shape of the human vocal tract during speech production~\cite{Principles_VoiceProduction} and are also affected by prosody~\cite{ProsodyEffectOnFormant}.
F1 is the lowest frequency formant, followed by F2, F3, and so on.
Typically, F1 and F2 suffice for vowel identification~\cite{SociophoneticsIntro}.
However, F3 adds an important layer of detail that enhances the precision of vowel identification~\cite{smit2010deviation}, aids in consonant distinction and provides critical information for speaker identification~\cite{almaadeed2016text} and speech intelligibility~\cite{amano2014determining}.
We compute the formant loss $L_\text{formant}$ as shown in Equation~\ref{eq:form}, where $\Phi^k(.)$ represents the $k^{th}$ formant. Here, $K$ and $N$ denote the total number of formants and frames, respectively.
\begin{equation}
\label{eq:form}
\begin{aligned}
    L_{\text{formant}} = \frac{1}{K \times N} \sum_{k=1}^{K} \sum_{n=1}^{N} (\Phi^k({x}_{n})
  - \Phi^k({x}_{dec_{n}}))^2
\end{aligned}
\end{equation}


\textbf{Representation-Driven Losses}:
Intermediate representations from self-supervised deep learning models capture a wide range of speech features, such as tonal quality, prosody, clarity, and background noise~\cite{pasad2021layer}, which are vital for assessing speech quality. Different network layers capture varying levels of abstraction, from basic acoustic features to more abstract representations like phonemes~\cite{nagamine2015exploring}.
Embeddings from supervised models trained for quality-related tasks offer richer information than standard element-wise loss functions~\cite{ghosh2021perception}.
Consequently, we compute the quality discrepancy as shown in Equation~\ref{equation: net_loss}, which is a general representation-driven loss, calculated on the activations ${\alpha}^j$ from the $j^{th}$ layer of a quality-based perceptual network.

\begin{equation} \label{equation: net_loss}
    L_{\text{DL}} = \frac{1}{\vert J \vert} \sum_{j\in J} \frac{1}{N}\sum_{n=1}^{N}(\alpha^j(x_{n}) - \alpha^j(x_{dec{_n}}))^2
\end{equation}

We consider two kinds of representation-driven losses:
\begin{enumerate}
[leftmargin=0.7cm,itemsep=0mm,labelsep=0.1cm,partopsep=0pt,topsep=0pt,parsep=0.5mm]
\item \textbf{Mean Opinion Score (MOS) Loss:}
The MOS is a widely used subjective metric for assessing the quality or naturalness of speech~\cite{Streijl2016MeanOS}.
However, incorporating human annotators to rate speech conversion during the training process is impractical.
To address this, we use a neural network, $Net_{\text{mos}}$, as a proxy for human evaluation, which is trained to predict the MOS score of a speech audio signal.
Specifically, we employ the model proposed in~\cite{Andreev_2023}, which consists of a fine-tuned Wav2Vec2.0 model~\cite{Wav2Vec} with a regression head added to the encoded features, resulting in a total of $\vert J \vert=4$ layer activations. The corresponding loss function $L_{\text{DL=mos}}$ is defined in Equation~\ref{equation: net_loss}, where the activations $\alpha$ are produced by the $Net_{\text{mos}}$ model.
\item \textbf{WavLM Loss:} WavLM~\cite{WavLM} is a state-of-the-art model for comprehensive speech processing tasks, demonstrating leading performance on SUPERB benchmarks~\cite{superb} in areas such as speaker verification and diarization. Studies like~\cite{KnnVC, GhoshAnonymising} highlight WavLM's capability to extract meaningful phoneme embeddings, with similar-sounding phonemes clustering in its latent space. The later layers of WavLM show reduced predictive power for pitch and prosody~\cite{UtilitySSLModels}, while the embeddings from layer $J$=6 are highly correlated with phoneme identification~\cite{KnnVC}. Therefore, we compute the WavLM loss $L_{\text{DL=wavlm}}$ using the activations from the 6\textsuperscript{th} layer, resulting in$\vert J \vert=1$ in Equation~\ref{equation: net_loss}.
\end{enumerate}

\textbf{Training Objectives}:
Put together, the full objective function of our proposed approach consists of the following terms that are weighted by $\lambda_i$, where $i \in \{\text{recon}, \text{code}, \text{comm}, \text{mos}, \text{wavlm}, \text{formant}\}$:
\begin{equation}
\hspace{-0.8em}
    \begin{aligned}
        L &= \lambda_{\text{recon}} L_{\text{recon}} + \lambda_{\text{code}} L_{\text{code}} + \lambda_{\text{com}} L_{\text{com}} + \lambda_{\text{mos}} L_{\text{DL=mos}} + \lambda_{\text{wavlm}} L_{\text{DL=wavlm}} + \lambda_{\text{formant}} L_{\text{formant}}
    \end{aligned}
\end{equation}

\section{Experiment Details}
We use three datasets: VCTK~\cite{VCTK_Dataset} and LibriSpeech~\cite{LibriSpeech_Dataset} for English utterances, and mlsGerman~\cite{Multilingual_LibriSpeech_Dataset} for German. Utterances are re-sampled to 16 kHz. The vanilla hierarchical VQVAE without perception-informed losses serves as our baseline model ($M_{\text{base}}$), while our proposed model ($M_{\text{PL}}$) incorporates these losses. All models are trained on the same splits and evaluated on the same test set. Log mel-spectrograms are used as input to the models. The models are optimized using the Adam optimizer with a cyclic learning rate, ranging from \(5 \times 10^{-4}\) to \(2 \times 10^{-3}\). The models are trained from scratch, employing early stopping with predicted mean opinion score (pMOS)~\cite{Andreev_2023} on the validation set as the stopping criterion. A pre-trained HiFiGAN vocoder~\cite{EmoStarGAN} is used to generate the waveform from the model's output. Additional details are provided in the appendix.

\subsection{Evaluation} 
We evaluate the baseline model ($M_{\text{base}}$) and our approach ($M_{\text{PL}}$) across three scenarios using both objective and subjective measures:
\begin{enumerate}[leftmargin=0.7cm,itemsep=0mm,labelsep=0.1cm,partopsep=0pt,topsep=0pt,parsep=0.5mm]
    \item \textbf{English$\rightarrow$English}: Both source and target speakers are English-speaking, evaluated within the same corpus (VCTK$\rightarrow$VCTK) and across different corpora (LibriSpeech$\rightarrow$VCTK). We also assess inter-accent conversion (Canadian, American, British).
    \item \textbf{German$\rightarrow$ English}: German utterances are converted using English VCTK target speakers.
    \item \textbf{German$\rightarrow$ German}: Both source and target speakers are German, using the mlsGerman dataset.
\end{enumerate}
In each scenario, we have 10 source speakers, each providing 10 utterances, and 10 target speakers.
The speakers are selected randomly, ensuring a disjoint set and balanced gender distribution, leading to 1000 total conversions.

\textbf{Objective Measures:} Intelligibility is measured by character error rate (CER) using the transcriptions from Whisper~\cite{Whisper} \textit{medium-english} model for the English conversions and \textit{medium} model for German conversions.
For anonymization, we measure the equal error rate (EER) using the speaker verification model ECAPA-TDNN~\cite{ECAPATDNN}, as in \cite{EmoStarGAN}.
We avoid using predicted MOS (pMOS) for quality evaluation, as in ~\cite{EmoStarGAN}, as it is used as a loss function in our model and could lead to overfitting, as discussed in ~\cite{de2024pesqetarian}. Instead, we rely on subjective testing for quality assessment.

\textbf{Subjective Evaluation Setup:} We evaluated quality, prosody preservation, intelligibility and anonymization by user studies via the Crowdee platform\footnote{https://www.crowdee.com/}.
We evaluated a random selection of 100 conversions for each of the three scenarios, as assessing all conversions would be both time-consuming and costly.
72 online participants had taken part in the studies.
For English$\rightarrow$English and German$\rightarrow$German scenarios, only native speakers of English and German, respectively, were allowed to participate.
In the German$\rightarrow$VCTK (English) scenario, native German speakers who were proficient in English were considered.
Participants rated quality (naturalness) on a scale from 1 (poor) to 5 (excellent).
They compared intonation and stress patterns between the original and converted samples for prosody preservation.
Intelligibility was assessed by selecting the most intelligible sample between the two conversions, for the same source utterance.
Anonymization was evaluated by rating the similarity on a scale ranging from 1 (different) to 5 (similar), between a converted sample and another utterance from the same speaker.
Each task was rated by at least 3 subjects, who were unaware of whether the samples were original or converted.
Trap questions and anchoring examples were used to ensure accuracy, and raters who failed trap questions twice were excluded.

\section{Results and Discussion}
\begin{figure*}
    \centering
    \includegraphics[width=\textwidth]{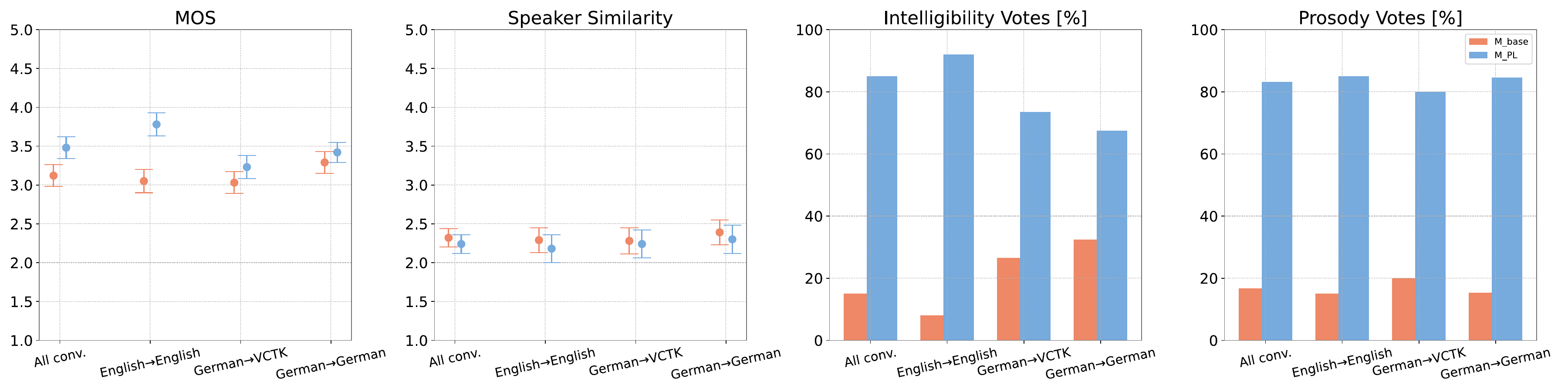}
    \caption{User study results for different scenarios and all conversions (All conv.). The Speaker Similarity plot indicates the similarity between the source and converted utterances (lower is better). The MOS plot shows the naturalness ratings from the user study (higher is better). The Prosody and Intelligibility Votes plots show the percentage of votes each model received. The mean MOS of the original files is 3.54.}
    \label{fig:subjective}
    \vspace{-1em}
\end{figure*}

Overall, our proposed method $M_{\text{PL}}$, significantly enhances intelligibility compared to the baseline $M_{\text{base}}$, as demonstrated in Table~\ref{tab:Metrics}. This improvement is also accompanied by improvement in naturalness corroborated by the MOS ratings from user studies, as shown in Figure~\ref{fig:subjective}.
Additionally, 83\% of the conversions using the proposed model were rated as more intelligible than those from the baseline.
\begin{table*}[ht]
\caption{Objective evaluation results are presented with 95\% confidence intervals.}
\label{tab:Metrics}
\centering
\resizebox{\textwidth}{!}
{
 \begin{tabular}{l|l|ll|ll}
 \toprule
 {\textbf{\specialcell{Source}}} & {\textbf{\specialcell{All conversions (All) \\ /Accent-wise}}} &  \multicolumn{2}{c|}{\textbf{CER} [\%] $\downarrow$}  & \multicolumn{2}{c}{\textbf{EER} [\%] $\uparrow$} \\
\cline{3-6}
{} & {} & {\textbf{$M_{\text{base}}$}} & {\textbf{$M_{\text{PL}}$}} & {\textbf{$M_{\text{base}}$}} & {\textbf{$M_{\text{PL}}$}}  \\ 
\midrule
{\textbf{All Conversions}} & {All}  & 71.33$\pm$0.40 & \textbf{53.09}$\pm$ 0.67  &41.07 & \textbf{43.21}\\
 \midrule
 \multirow{6}{*}{\shortstack{\textbf{VCTK} $\rightarrow$ \textbf{VCTK}}}  & {All}  & 73.32$\pm$0.79  & \textbf{45.49}$\pm$1.13  & 37.88 & \textbf{38.17 }
\\
 \cmidrule{2-6}
 & {American $\rightarrow$ British}   &  73. 31$\pm$1.49   & \textbf{45.32}$\pm$2.24 & 36.89 & \textbf{39.10}\\ 
 & {Canadian $\rightarrow$ British}  &  72.02$\pm$1.39  & \textbf{48.89}$\pm$1.97  & 37.10 & \textbf{42.23}
\\
 & {British $\rightarrow$ British}  &  74.30$\pm$1.26  & \textbf{43.07}$\pm$1.68    & 36.07 & \textbf{38.02}
\\
 \midrule
 \multirow{1}{*}{\textbf{\shortstack{\textbf{LibriSpeech} $\rightarrow$  \textbf{VCTK}}}}  & {All} & 70.16$\pm$0.64  & \textbf{53.50}$\pm$1.12  & \textbf{38.26} & 38.00
\\
\midrule
\multirow{1}{*}{\textbf{\shortstack{\textbf{German} $\rightarrow$ \textbf{VCTK}}}}  
& {All}  & 77.44$\pm$0.56  & \textbf{75.21}$\pm$0.75  & 42.35 & \textbf{44.84}
\\
\midrule
\multirow{1}{*}{\textbf{\shortstack{\textbf{German} $\rightarrow$ \textbf{German}}}}  & {All} & 64.41$\pm$0.91  & \textbf{38.15}$\pm$1.02  & 49.63 & \textbf{51.46} \\
 \bottomrule
\end{tabular}
}
\end{table*}
In terms of speaker anonymization, there is a modest increase in EER from 41.07\% to 43.21\% across all scenarios, which is similarly reflected in the speaker similarity scores from user studies. For prosody preservation, our approach significantly outperforms the baseline, with 83.2\% of participants favouring the proposed model having perception-informed losses, as seen in Figure~\ref{fig:subjective}.
Similar trends are observed for within-corpus scenario VCTK$\rightarrow$VCTK, where the mean CER showed a significant improvement from 73.32\% to 45.49\% with the incorporation of the proposed losses, as detailed in Table~\ref{tab:Metrics}.
These improvements are also observed in cross-gender (refer to Appendix) and cross-accent conversions within the corpus.
For prosody preservation and intelligibility, our model received significantly higher support with 83\% and 85\% of the votes, respectively. Furthermore, in inter-accent conversions, we observe a change in accent after the VC, where the converted sample adopted the accent of the target speaker, potentially leading to better anonymization.
In the cross-corpus scenario LibriSpeech$\rightarrow$VCTK, similar trends are observed for all metrics.

In the German to English (VCTK) conversion, intelligibility did not improve much compared to intra-lingual conversions, as shown in Table~\ref{tab:Metrics}. Listening to samples\footnote{Audio samples available at: \url{https://shorturl.at/mqSrs}} reveals that using English target speakers introduced an English accent in the conversions, failing to preserve the original intonation.
This occurred because German phonemes that do not exist in English were replaced by similar English sounds.
For example, the German uvular fricative [\textipa{\textinvscr}] in ``Rad'' became the alveolar \textipa{[\textturnr]} as in ``run''. The German fricatives \textipa{[ç]} (``ich'') and \textipa{[x]} (``ach'') were replaced by \textipa{[\textesh]} (``sh'') and \textipa{[k]} (``cat''). This likely occurs because the VQVAE model is trained on English data, substituting German sounds with the closest English equivalents. However, this accent shift aids anonymization, potentially leading to a higher EER compared to the VCTK$\rightarrow$VCTK scenario.
For German$\rightarrow$German conversions, a similar improvement is observed as in the English$\rightarrow$English scenario. However, regarding naturalness, $M_{\text{PL}}$ shows less improvement compared to the English$\rightarrow$English scenario, as indicated in the user study results (MOS score in Figure~\ref{fig:subjective}). This might be attributed to the formant prediction network $Net_{\text{formant}}$ being trained solely on English data. Consequently, the network may not accurately capture German vowel nuances, leading to a mismatch in vowel prediction that results in converted German speech sounding less authentic, as reflected in subjective evaluations.

\section{Conclusion}
We present model-agnostic perception-informed losses as an innovative approach to enhance the quality of voice conversion (VC) for speech anonymization without increasing model complexity. By integrating quality-related knowledge into the training process through handcrafted acoustic features and deep learning representations, our framework significantly improves the performance of a vanilla hierarchical VQVAE-based model. Augmented solely by our proposed loss functions, the model shows notable enhancements in naturalness, intelligibility, and prosody preservation across diverse conversion scenarios, including cross-corpus conversions, varying genders, accents, and languages. Objective and subjective evaluations validate these results, highlighting the importance of incorporating speech-specific features within the loss function, rather than increasing model complexity. Looking ahead, we plan to develop loss functions to specifically target and reduce the graininess observed in some conversions.
\section{Acknowledgements}
This research has been supported by the Federal Ministry of Education and Research of Germany through project Emonymous (project number S21060A) and Medinym (focused on AI-based anonymization of personal patient data in clinical text and voice datasets).
\bibliographystyle{IEEEtran}
\bibliography{perceptual_loss}

\newpage
\section{Appendix}
\subsection{Detailed Objective Evaluation}
Table~\ref{tab:Metrics-det} shows a more detailed version of the objective evaluation portrayed in Table~\ref{tab:Metrics}, where gender-wise information is also mentioned.
\begin{table*}[h]
\caption{Objective evaluation results are presented with 95\% confidence intervals.}
\label{tab:Metrics-det}
\centering
\resizebox{\textwidth}{!}
{
 \begin{tabular}{l|l|ll|ll}
 \toprule
 {\textbf{\specialcell{Source}}} & {\textbf{\specialcell{All conversions (All) \\ /Accent-wise}}} &  \multicolumn{2}{c|}{\textbf{CER} [\%] $\downarrow$}  & \multicolumn{2}{c}{\textbf{EER} [\%] $\uparrow$} \\
\cline{3-6}
{} & {} & {\textbf{$M_{\text{base}}$}} & {\textbf{$M_{\text{PL}}$}} & {\textbf{$M_{\text{base}}$}} & {\textbf{$M_{\text{PL}}$}}  \\ 
\midrule
{\textbf{All Conversions}} & {All}  & 71.33$\pm$0.40 & \textbf{53.09}$\pm$ 0.67  &41.07 & \textbf{43.21}\\
 \midrule
 \multirow{6}{*}{\shortstack{\textbf{VCTK} $\rightarrow$ \textbf{VCTK}}}  & {All}  & 73.32$\pm$0.79  & \textbf{45.49}$\pm$1.13  & 37.88 & \textbf{38.17 }
\\
 \cmidrule{2-6}
 & {American $\rightarrow$ British}   &  73. 31$\pm$1.49   & \textbf{45.32}$\pm$2.24 & 36.89 & \textbf{39.10}\\ 
 & {Canadian $\rightarrow$ British}  &  72.02$\pm$1.39  & \textbf{48.89}$\pm$1.97  & 37.10 & \textbf{42.23}
\\
 & {British $\rightarrow$ British}  &  74.30$\pm$1.26  & \textbf{43.07}$\pm$1.68    & 36.07 & \textbf{38.02}
\\
\cmidrule{2-6}
  & {Different gender}   & 73.05$\pm$1.12  & \textbf{46.24}$\pm$1.58  & - & -  \\
& {Same gender} & 73.60$\pm$1.12  & \textbf{44.72}$\pm$1.61   & - & - \\
 \midrule
 \multirow{1}{*}{\textbf{\shortstack{\textbf{LibriSpeech} $\rightarrow$  \textbf{VCTK}}}}  & {All} & 70.16$\pm$0.64  & \textbf{53.50}$\pm$1.12  & \textbf{38.26} & 38.00
\\
\cmidrule{2-6}
& {Different gender} & 69.71$\pm$0.92  & \textbf{53.73}$\pm$1.61   & - & -
\\
& {Same gender} & 70.63$\pm$0.88  & \textbf{53.27}$\pm$1.58   & - & -
\\
\midrule
\multirow{1}{*}{\textbf{\shortstack{\textbf{German} $\rightarrow$ \textbf{VCTK}}}}  
& {All}  & 77.44$\pm$0.56  & \textbf{75.21}$\pm$0.75  & 42.35 & \textbf{44.84}
\\
\cmidrule{2-6}
& {Different gender} & 76.93$\pm$0.79  & \textbf{75.71}$\pm$1.03  &- & -
\\
& {Same gender} & 77.96$\pm$0.80  & \textbf{74.70}$\pm$1.08  & - & -
\\
\midrule
\multirow{1}{*}{\textbf{\shortstack{\textbf{German} $\rightarrow$ \textbf{German}}}}  & {All} & 64.41$\pm$0.91  & \textbf{38.15}$\pm$1.02  & 49.63 & \textbf{51.46} \\
\cmidrule{2-6}
& {Different gender} & 65.87$\pm$1.27  & \textbf{39.17}$\pm$1.44  & - & -
\\
& {Same gender} & 63.00$\pm$1.31  & \textbf{37.17}$\pm$1.45   & - & - \\
 \bottomrule
\end{tabular}
}
\end{table*}
\subsection{Ablation Study}

We perform ablation studies to assess the contribution of each loss component. Table~\ref{tab:abl} demonstrates that formant $L_{\text{formant}}$ individually contributes the most to naturalness and intelligibility. This suggests that calculating loss on specific frequency components effectively enhances the overall quality of VC. These components correspond to the resonant frequencies of the vocal tract, which are essential for perceiving vowel sounds and overall intelligibility. Further, listening to the samples reveals that the model not trained with $L_{\text{formant}}$ has the worst prosody preservation.
Removing $L_{\text{formant}}$ loss (when using $L_{\text{DL=wavlm}}$ + $L_{\text{DL=mos}}$) significantly increases the CER from 50.02\% to 66.85\%, highlighting the critical role of formants in speech intelligibility.
\begin{table}[!h]
\centering
\caption{Ablation study results with 95\% confidence intervals shown on the VCTK $\rightarrow$ VCTK conversion setup. pMOS is the MOS score predicted by Net\textsubscript{mos}.}

\label{tab:abl}
\begin{tabular}{l| c c c c c}

\toprule
\textbf{Method} & {\textbf{pMOS} $\uparrow$}  & {\textbf{CER} [\%] $\downarrow$} & {\textbf{EER} [\%] $\uparrow$} \\
\midrule
$M_{\text{base}}$ & {1.59 $\pm$0.04} & {73.32$\pm$0.79}  & 37.88  \\
$M_{\text{PL}}$ &  {\textbf{3.56} $\pm$0.02}  & {\textbf{45.49} $\pm$1.13} & \textbf{38.17}  \\
\midrule
{$L_{\text{formant}}$} & {\textbf{3.13} $\pm$0.01}  & {\textbf{50.67} $\pm$1.02}   & 37.07  \\
{$L_{\text{DL=wavlm}}$} & {3.02 $\pm$0.02}  & {51.74 $\pm$3.02}   & 37.89  \\
{$L_{\text{DL=mos}}$} & {2.33 $\pm$0.04}  & {68.17 $\pm$2.02}   & \textbf{37.97}  \\
\midrule
{$L_{\text{formant}}$ + $L_{\text{DL=mos}}$} & {2.71 $\pm$0.02}  & {50.02 $\pm$1.02}   & \textbf{38.16}  \\
{$L_{\text{formant}}$ + $L_{\text{DL=wavlm}}$} & {\textbf{3.47} $\pm$0.01}  & {\textbf{49.28} $\pm$1.08}& 38.12  \\
{$L_{\text{DL=wavlm}}$ + $L_{\text{DL=mos}}$} & {2.34 $\pm$0.03}  & {66.85 $\pm$1.02}   & 37.91  \\

\bottomrule
\end{tabular}

\end{table}
\raggedbottom

Interestingly, individually $L_{\text{formant}}$ and $L_{\text{DL=wavlm}}$ have a greater positive impact on MOS scores compared to $L_{\text{DL=mos}}$, indicating that these losses better capture the aspects of speech that influence perceived quality. One reason $L_{\text{DL=mos}}$ underperforms compared to $L_{\text{DL=wavlm}}$ is that WavLM was trained on a much larger corpus to capture more generic speech representations, encompassing noise, distortion, natural variations in pitch, loudness, and other factors. In contrast, the MOS network is specifically trained to predict MOS scores, focusing solely on naturalness.

The combination of $L_{\text{formant}}$ and $L_{\text{DL=wavlm}}$ significantly improves the pMOS and CER compared to the baseline, nearly reaching the performance of the model incorporating all losses $M_{\text{PL}}$.
We also note that the anonymization capability of the model is not significantly affected by the removal of the loss components individually or in combination. This indicates that the mechanisms responsible for anonymization are robust and independent of the specific losses used to enhance naturalness and intelligibility.

\subsection{Training Details}

We trained all models on log mel-spectrograms with 80 mel bands, generated from 2-second audio clips. For STFT parameters, we used a hop length of 320 and a window length of 1024.

For scenarios involving English-speaking target speakers, our models were trained on approximately 5 hours of English utterances from 20 speakers in the VCTK dataset, with the data divided into a 90:10 split for training and validation. In cases requiring German-speaking targets, we utilized around 10 hours of German utterances from 20 speakers in the mlsGerman dataset, allocating 80\% for training and 20\% for validation.

The number of trainable parameters in all the voice conversion models is the same, as we only augment the vanilla model with our proposed losses. Training with all three perception-aware losses required approximately 2 days on average to complete on a 80\SI{}{\giga\byte} A100 GPU. We set $\lambda_{\text{recon}} = 1$, $\lambda_{\text{code}} = 1$, $\lambda_{\text{comm}} = 3$, $\lambda_{\text{mos}} = 0.1$, $\lambda_{\text{wavlm}} = 0.1$, and $\lambda_{\text{formant}} = 10^6$, ensuring that all loss terms were within the same order of magnitude. We incorporated $L_{\text{DL=mos}}$ from epoch 0 and $L_{\text{DL=wavlm}}$, $L_{\text{formant}}$ from epoch 45 into the training based on empirical observations obtained during the development phase.

We used a pre-trained HiFiGAN~\cite{HiFi_GAN} vocoder from~\cite{EmoStarGAN} to generate the waveform from the mel-spectrogram, which produced a one-minute long waveform from the converted mel-spectrogram in 0.1 seconds on the A100.

We trained the $Net_{\text{formant}}$ model to derive the F1, F2, and F3 values needed to compute the $L_{\text{formant}}$ loss. The formant network consists of a transformer encoder architecture as proposed in~\cite{AttentionIsAllYouNeed}, additionally featuring a regression head with three output neurons that predict based on the encodings of the input for each time frame. As training data, we used the VTR dataset~\cite{VTR_Dataset}, comprising 538 manually formant-annotated utterances from domain experts who ensured balance across phonetic contexts, speakers, genders, and dialects in the English language. We used the default VTR parameters for pre-processing and achieved a final MSE of 3.16.
\end{document}